# 유전적 랜덤 가중치 변화 학습 알고리즘 기반 신경회로망 최적화 기법

## Optimizing method for Neural Network based on Genetic Random Weight Change Learning Algorithm

○Mohammad Ibrahim sarker [1], Zubaer Ibna Mannan [2], Hyongsuk Kim [3]
[1] 전북대학교 전자공학부 (TEL: 063-270-2477; E-mail: sarkeribrahim@gmail.com)
[2] 전북대학교 전자공학부 (TEL: 063-270-2477; E-mail: zimannan@gmail.com)
[3] 전북대학교 전자공학부, 지능형로봇연구센터 (TEL: 063-270-2477; E-mail: hskim@jbnu.ac.kr)

**Abstract** Random weight change (RWC) algorithm is extremely component and robust for the hardware implementation of neural networks. RWC and Genetic algorithm (GA) are well known methodologies used for optimizing and learning the neural network (NN). Individually, each of these two algorithms has its strength and weakness along with separate objectives. However, recently, researchers combine these two algorithms for better learning and optimization of NN. In this paper, we proposed a methodology by combining the RWC and GA, namely Genetic Random Weight Change (GRWC), as well as demonstrate a seminal way to reduce the complexity of the neural network by removing weak weights of GRWC. In contrast to RWC and GA, GRWC contains an effective optimization procedure which is worthy at exploring a large and complex space in intellectual strategies influenced by the GA/RWC synergy. The learning behavior of the proposed algorithm was tested on MNIST dataset and it was able to prove its performance.

## 1. Introduction

Implementation of multilayer neural network in hardware has long been a problem for the scientists and researchers. The complexity of the learning algorithms is the vital problem to implement the NNs in hardware. However, some researchers have fabricated on-chip learning neural networks. Arima et al. implemented a revised Boltzmann Machine learning algorithm with analog/digital hybrid circuitry [1], Morie et al. fabricated an analog neural network chip with Back Propagation algorithm [2]. Shima et al. [3] employed digital circuitry for implementation of Back Propagation. These traditional learning algorithms are expressed with complex equations and require complicated multiplication. Unlike these traditional algorithms, random weight change algorithm is easy and suitable for analog implementation even though random weight change algorithm is less efficient and undergoes local minima problems than back propagation.

In contrast to RWC, Genetic algorithm are inspired by Darwin's theory about evolution- "survival of the fittest". Wang et al. [4] used genetic algorithm (GA) with artificial neural network (ANN) to find out optimal process parameters for optimal performances.

Combining different algorithms can fetch better results than they could achieve individually. In this paper, we propose a modification of random weight change (RWC) algorithm which allows us to combine genetic algorithm (GA) with random weight change (RWC) algorithm, we called it as genetic random weight change algorithm (GRWC). The results obtained through this method shows an improvement than the conventional random weight change method.

## 2. Proposed Algorithm

The combining process is applied on a function, where genetic algorithm is used to set parameters for training the random weight change algorithm. In nature world, we cannot say an individual is strong or not when it is newly born. We should let it develop for a period of time and make the decision whether to let it breed or eliminate it. Therefore, in our proposed algorithm, eight simulations run's parallelly for 1000 epochs, then choose the best two and let them copy and reproduce. Our Genetic random weight change algorithm operates according to following steps

Step of algorithm:

1) In Initialization part, a number of candidate weights of the neural network is generated, very often these have random values

$$\Theta_i^{(1)} = \delta \cdot rand, \quad \Theta_i^{(2)} = \delta \cdot rand, \quad i = 1, 2, \cdots, N$$
$$\Delta\Theta_i^{(1)} = \delta \cdot rand, \quad \Delta\Theta_i^{(2)} = \delta \cdot rand, \quad i = 1, 2, \cdots, N$$

2) Calculate cost function J for each candidate called Evaluation. A cost function will allow to score the prediction performance

※ This work was carried out with the support of " Development of real-time diagnosis analysis technology for the major disease and insects of tomato using neural networks (Project No. PJ0120642017)", NRF (project No.-2016R1A2B4015514 BK 21 Plus and Intelligent Robot Research Center, Chonbuk National University, Republic of Korea.

of each candidate; the score will be a number that tells how good this solution solves the problem.

$$z_{2,i} = f(\Theta_i^{(1)} \cdot x), \quad z_{3,i} = f(\Theta_2^{(2)} \cdot z_{2,i})$$

$$h_{\Theta,i}^j = \frac{z_{3,i}^j}{\sum_{j=1}^{N\_label} z_{3,i}^j}, \quad J_i = \frac{1}{2}(h_{\Theta,i} - y)^2$$

where, $i = 1, 2, \ldots, N$.

Selection: select two best candidates from the groups according to the cost value each individual can achieve.

$$Index_{1,2} = \min(J)$$

3) Update neural network weight:

Copy the neural weight from the best two selected candidates. Abandon the value of other candidates and assign their weights from the best 2 candidate.

$$\Theta_i = \Theta_{Index_1}, \ \Delta\Theta_i = \Delta\Theta_{Index_1} \quad i=1,2,\cdots,n/2$$
$$\Theta_i = \Theta_{Index_2}, \ \Delta\Theta_i = \Delta\Theta_{Index_2} \quad i=n/2+1, n/2+2, \cdots, n$$

With a mutation based on random weight change algorithm of all the candidates, we get the offspring weight and the cost value of each individual.

$$\Theta_i^{(1)}(n+1) = \Theta_i^{(1)}(n) + \Delta\Theta_i^{(1)}(n+1)$$

where

$$\Delta\Theta_i^{(1)}(n+1) = \Delta\Theta_i^{(1)}(n) \text{ if } J_i(n+1) < J_i(n)$$
$$\Delta\Theta_i^{(1)}(n+1) = \lambda \cdot rand \text{ if } J_i(n+1) \geq J_i(n)$$

Test the cost function value and prediction correction based on test data.

4) By repeating following steps until the cost goes down below satisfactory value.

For reducing the complexity of hardware implementation, we remove all the weak weights (lower magnitude weights) after the threshold value 0.1. In hardware implementation of NN, less numbers of weights mean less number of connections are needed to implement a successive neuronal communication. In our experiment, we were able to remove almost half of the total weights of the NN. The general idea of proposed algorithm is illustrated in this figure 1.

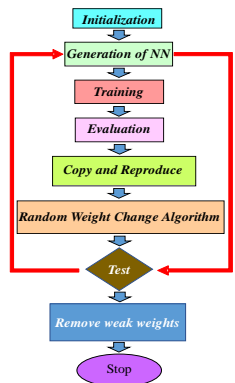

Fig. 1. The Principle Structure of Proposed Algorithm.

## 3. Results

We trained our proposed network with 1000 images from MNIST dataset for handwritten character recognition problem also compared both RWC and GRWC. Figure 2 illustrate the average error curves for both RWC and GRWC algorithm. It is distinctly observable that the average error per iterations for GRWC is less than RWC algorithm. Figure 3 illustrates the average error curve of the proposed method where the lower amplitude weights are removed from the neural network with a threshold value of error 0.1. Scrutinizing of Fig. 3 illustrate that the error increases after the error value 0.1 as lower magnitude weights are removed from the network and the proposed network starts to perform fine tuning again. On the basis of error curve, the average error of GRWC (Fig. 2) is less than the proposed model (Fig. 3). However, the number of connections of NN presented in Fig. 2 is more than twice than the number of connections of NN presented in Fig. 3. This feature of the propose algorithm makes it extremely competent and robust for the hardware designing application.

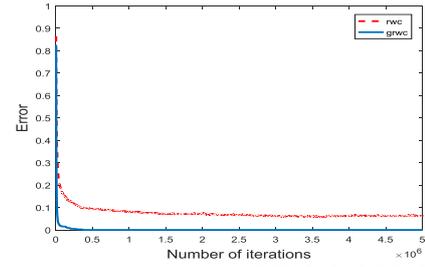

Fig. 2. Average error curve for RWC & GRWC.

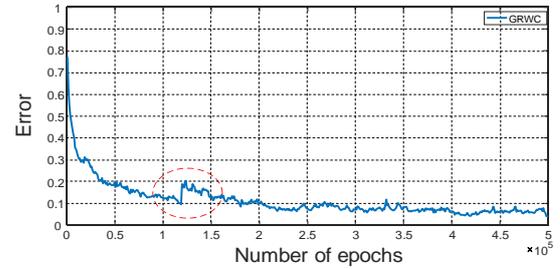

Fig. 3. Average error curve of proposed algorithm.

## 4. CONCLUSION

In this paper, we presented the GRWC algorithm where the effective learning of GRWC depends on the removing of weak weights from the neural networks. Through our proposed new GRWC algorithms we get seminal accuracy. However, the vital achievement of this study is that we reduced the size of our proposed network by half of the total size of the network as we remove the lower amplitude weights from the NN. This aspect of our proposed model insures that the model can easily be implemented in hardware design due to the less number of connections are required to implement this network in hardware design.